\documentclass[runningheads,a4paper]{llncs}

\usepackage{amssymb}
\usepackage{graphicx}
\usepackage{url}
\usepackage{bibnames}
\usepackage{wrapfig}
\usepackage{url}
\urldef{\mailsa}\path|{rw37, pg427, grs53}@drexel.edu|

\urldef{\mailsc}\path| |    
\newcommand{\keywords}[1]{\par\addvspace\baselineskip
\noindent\keywordname\enspace\ignorespaces#1}

\begin{document}

\mainmatter  

\title{Longitudinal Distance: \\Towards Accountable Instance Attribution}

\author{Rosina O. Weber\textsuperscript{\rm 1},
        Prateek Goel\textsuperscript{\rm 1},
        Shideh Amiri\textsuperscript{\rm 3}, and 
        Gideon Simpson\textsuperscript{\rm 4}
}

\authorrunning{Weber et al.}
\institute{Information Science, Department of Mathematics, Drexel University\\ 
{\{rw37, pg427, ss4588, grs53\}}@drexel.edu
}

\maketitle

\begin{abstract}
Previous research in interpretable machine learning (IML) and explainable artificial intelligence (XAI) can be broadly categorized as either focusing on seeking interpretability in the agent’s model (\emph{i.e}., IML) or focusing on the context of the user in addition to the model (\emph{i.e}., XAI). The former can be categorized as feature or instance attribution. Example- or sample-based methods such as those using or inspired by case-based reasoning (CBR) rely on various approaches to select instances that are not necessarily attributing instances responsible for an agent’s decision. Furthermore, existing approaches have focused on interpretability and explainability but fall short when it comes to accountability. Inspired in case-based reasoning principles, this paper introduces a pseudo-metric we call Longitudinal distance and its use to attribute instances to a neural network agent’s decision that can be potentially used to build accountable CBR agents.

\end{abstract}

\setlength{\floatsep}{12.0pt} 
\setlength{\textfloatsep}{10.0pt} 
\setlength{\intextsep}{6.0pt} 
\setlength{\abovecaptionskip}{6.0pt} 
\setlength{\belowcaptionskip}{6.0pt} 

\section{Introduction and Background}
The literature in interpretable machine learning (IML) and explainable artificial intelligence (XAI) have proposed various categorization schemes and taxonomies to describe their methods. One of those is through attribution -- attribution methods attempt to explain model behavior by associating a classified (or predicted) test instance to elements of the model 
(\emph{e.g}., \cite{olah2018building,sundararajan2017axiomatic}). 

There are two main categories of attribution methods. Instance attribution methods select instances as integral elements to associate with a classification or prediction (\emph{e.g}., \cite{barshan2020relatif,khanna2019interpreting,koh2017understanding,mercier2020interpreting,yeh2018representer,sliwinski2019axiomatic}). Instance attribution methods have been used for debugging models, detecting data set errors, and creating visually-indistinguishable adversarial training examples \cite{barshan2020relatif,koh2017understanding,mercier2020interpreting,yeh2018representer}. Feature attribution methods select features to attribute to a classification or prediction indicating which features play a more significant role than others \cite{bach2015pixel,bhatt2019building,lundberg2017unified,ribeiro2016should,shrikumar2017learning,sundararajan2017axiomatic}.

When using case-based reasoning (CBR), a third category is often mentioned: example- or sample-based. CBR can offer instances as examples but it does not make attributions. The most successful use of CBR for explainability or interpretability is in the ANN-CBR twin by \cite{kenny2019twin}. In their work, Kenny and Keane \cite{kenny2019twin} were able to transfer the accuracy of a CNN into CBR for full transparency. This was done by applying the feature attribution approach DeepLift \cite{shrikumar2017learning}. 

Both instance and feature attribution methods suffer strong criticisms. DeepLift \cite{shrikumar2017learning} demonstrated good performance when used in the investigation reported in \cite{kenny2019twin}. However, feature attribution methods have been shown to be insensitive to changes in models and in data \cite{adebayo2018sanity}. In addition, they do not work in deep learning (DL) models that use a memory ~\cite{koul2018learning}. One important limitation of feature attribution is how to evaluate them because of how the data distribution changes when a feature is removed \cite{NEURIPS2019_fe4b8556,adler2018auditing}.

If we extend the analysis beyond interpretability and explainability and consider accountability, then feature attribution methods pose more challenges. An accountable decision-making process should demonstrate its processes align with legal and policy standards \cite{kroll2017accountable,mittelstadt2019explaining}. This notion is expanded in the literature around two goals. The first is to generate a fully reproducible \cite{kroll2017accountable} audit of the system generating the decision (\emph{e.g}.,\cite{adler2018auditing,baldoni2016computational,mittelstadt2016ethics,now2018annual}); the second is to change the system when its decision is unwarranted (\emph{e.g}., \cite{garfinkel2017toward,mittelstadt2019explaining}). The additional challenge to feature attribution lies therefore on the difficulty to change the behavior of a system that produces unwanted decisions based on simply knowing the role of features in each decision. Notwithstanding, it seems plausible to make changes when the instances responsible for each decision are known. 

Instance attribution methods are also target of criticisms. One criticism has been directed to instances attributed when using influence functions \cite{koh2017understanding} by Brashan et al.  \cite{barshan2020relatif}. It refers to the fact that the selected instances are mostly outliers, a problem they overcome with relative influence \cite{barshan2020relatif}. Another criticism is how time and processing consuming it is to compute influence functions \cite{khanna2019interpreting}. A third criticism is that selected sets of instances have substantial overlap across multiple test instances to be classified. Consequently, there seems there is still much to gain from further examining instance attribution.

This paper proposes a new instance attribution method. It is intuitive and thus simple to explain to users. It is also simple to compute. It has limitations as well. It shares with influence functions the overlap between sets of instances attributed to multiple test instance. It shares with ROAR \cite{NEURIPS2019_fe4b8556} and HYDRA \cite{chen2021hydra} the requirement to retrain the model. It is a step into a different direction that may spur further research. The proposed method is based on a new pseudo-metric we call Longitudinal Distance (LD). This paper focuses exclusively on classifications. For simplicity, we henceforth refer to solutions as classifications and the NN structures as classifiers. In the next sections, we describe the pseudo-metric LD and a non-pseudo-metric variant and how they can be used for explanation. We then present some preliminary studies. We conclude with a discussion and future steps.

\section{Introducing Longitudinal Distances}
Longitudinal distances are a class of distances based on a heuristic that an iterative learning model such as NN can be used as an oracle for instance attribution. Longitudinal distances operate in the metric space of instance elements that are used to train and to be solved by NN methods through a series of iterations (or epochs).

Given a classification model trained with neural networks (NN) on a space where $x \in X$ are instances mapped by features $f \in F$, $X_{\rm train} \in X$ are training instances and hence include labels $y \in Y$ to indicate the class they belong, and $X_{\rm test}$ are testing instances. The classifier $C_e(x_i)$ is a NN classifier that learns to assign labels $y $ using a set of $n$ training instances $x_i$ through $e $ epochs, $e = 1,\ldots,k$, $x_i = 1,\ldots, n$.
We assume that the ability of a classifier to solve a previously unseen instance is represented in the weights learned through a sequence of learning iterations from a set of training instances. We therefore hypothesize that there is at least one (or more) training instance(s) responsible for the classification of an unseen instance, be it correct or incorrect, and that the relation between the training and unseen instances are somehow represented throughout the sequence of learning iterations.

We justify the proposed hypothesis based on the fact that when a previously known solution, {\it i.e.}, a known class, is selected to classify an unseen problem, that both the training instance and the unseen instance, now both labeled as members of the same class, meet some condition that causes their solutions to be interchangeable, that is, their labels are the same. This condition is an abstract concept that we wish to represent, let us call it {\it the oracle condition}. We know that two instances together met this oracle condition when they are members of the same class. Consequently, a trained classifier can be perceived as a function that indicates that two instances meet said oracle condition. However a trained classifier may not be ideal to distinguish, among all instances that meet the oracle condition, which ones are responsible for the classification of an unknown instance. We consider, for example, the problem known as \emph{catastrophic interference} \cite{ffe0793b43f842d2a50467d736a80c83} to possibly degrade the quality of a classifier with respect to some classes as it continues to learn new classes. Because some abilities may be forgotten as an NN method moves through new epochs, we contend that the complete experience learned by a classifier is better represented by every weight matrix learned in all epochs of its learning cycle. The last weight matrix represents the learned classifier that has refined its ability to classify all instances through generalization. Therefore, to successfully generalize, it is reasonable to assume that particular abilities with respect to some instances or labels were sacrificed. Hence, we propose to use the sequence of all intermediary classifiers, the weight matrices resultant at the end of each learning iteration, to measure the relationship between an unknown instance and all the training instances and identify the one(s) responsible for the classification. 

We first propose the pseudometric Longitudinal Distance, ${d_L}$ to produce the ratio of incorrect classifications to the total number of epochs. 
\begin{equation}
\label{e:dC}
    d_{L}(x_i,x) = 1 - ( \frac{1}{k}\sum_{e=1}^{e=k} \delta_{e}(x_i,x))
\end{equation}

Now we propose the non-pseudometric Strict Longitudinal Distance, 
\begin{equation}
\label{e:dCC}
    d_{SL}(x_i,x) = 1 - \left( \frac{\sum_{e=1}^{e=k} w_e(x_i) \delta_{e}(x_i,x)}{\sum_{e=1}^{e=k} w_e(x_i)}\right)
\end{equation}

We define the distance space as $(d,X)$ for the distance pseudo-metric $d_{L}(x_i, x)$, along with $d_{SL}(x_i, x)$ where $x_i, x \in X$  are instances mapped by features $f \in F$ that are classified by a classifier $C_e$ at epoch $e$ and receive a label $y \in Y$ to indicate their outcome class. 

In the above expressions,
\begin{equation}
\label{e:se}
    \delta_e(x_i,x) = 1_{C_{e}(x_i)=C_{e}(x)}
\end{equation}
the $1_{C_{e}(xi)=C_{e}(x)}$ in Equation \ref{e:se} is the indicator function taking the value of  1 when they are equal and 0 otherwise. 

For assigning relevance, we incorporate $w_e(x_i)$ as a binary weight to Equation 2 that indicates whether the classifier $C_{e}(x_i)$ is correct or wrong. When $w_e(x_i) = 1$, this means the label $y$ predicted by the classifier $C_{e}(x_i)$ for $x_i$ is equal to the label of the instance and thus it is correct, otherwise $w_e(x_i) = 0$. We observe that Equation 1 assumes $x_i$ as a training instance and therefore there is a label $y$ designated as its class and thus $w_e(x_i)$ can be computed. Note that only the correctness of the classification of the training instance is verified as the correct label of unseen instances is unknown. Consequently Equation 2 is suited for computing the distances between a training instance $x_i \in X_{\rm train}$ and a testing instance $x \in X_{\rm test}$, which is the goal of this paper. 

Note that we can rewrite 
\begin{equation}
d_{L}(x_i,x) = \frac{1}{k}\sum_{e=1}^{e=k} (1-\delta_e(x_i,x)) =
    \frac{1}{k}\sum_{e=1}^{e=k} 1_{C_{e}(x_i)\neq C_{e}(x)}
\end{equation}%
which is the mean number of classifier mismatches over all the epochs.

Next we demonstrate that $d_{L}$ a pseudo-metric. 
Recall the pseudo-metric properties:
\begin{enumerate}
    \item $d(x,x)=0$
    \item $d(x,y) = d(y,x)$
    \item $d(x,y)\leq d(x,z) + d(z,y)$
\end{enumerate}
What makes $d$ a pseudo-metric and not necessarily a metric is that in order to be a metric, if $d(x,y)=0$, then $x=y$.  Because of mapping into feature space to evaluate the classifier, there may exist predictors, $x_i$ and $x$, for which $x_i\neq x$, but $d(x_i,x)=0$.

For $d_L(x_i,x)$,
\begin{equation}
    d_L(x_i,x) = \frac{1}{k}\sum_{e=1}^{e=k} 1_{C_{e}(x_i)\neq C_{e}(x)} =  \frac{1}{k}\sum_{e=1}^{e=k} 0 = 0.
\end{equation}
The symmetry property is also obvious. For the triangle property, given $x$, $y$, and $z$, suppose $C_{e}(x) = C_e(y)$.  Then, 
\begin{equation}
    1_{C_{e}(x)\neq C_{e}(y)} =0 \leq 1_{C_{e}(x)\neq C_{e}(z)}  + 1_{C_{e}(z)\neq C_{e}(y)},
\end{equation}
regardless of what the two terms on the right are.  If, instead, $C_{e}(x) \neq C_{e}(y)$, then
\begin{equation}
     1_{C_{e}(x)\neq C_{e}(y)} =1
\end{equation}
Now we argue by contradiction. Suppose for some choice of $z$, 
\begin{equation}
    1_{C_{e}(x)\neq C_{e}(z)}  + 1_{C_{e}(z)\neq C_{e}(y)} = 0
\end{equation}
For this to be true, each of these two terms must be zero, so 
\begin{equation}
    C_{e}(x)= C_{e}(z) = C_{e}(y)
\end{equation}
But this contradicts the assumption that $C_{e}(x) \neq C_{e}(y)$.  Thus, $ 1_{C_{e}(x)\neq C_{e}(z)}  + 1_{C_{e}(z)\neq C_{e}(y)}\geq 1$, and 
\begin{equation}
    1_{C_{e}(x)\neq C_{e}(y)} \leq 1_{C_{e}(x)\neq C_{e}(z)}  + 1_{C_{e}(z)\neq C_{e}(y)} 
\end{equation}
Consequently the triangle equality holds for each epoch, and it must then hold over any finite sum of epochs:
\begin{equation}
    d_L(x,y) = \frac{1}{k}\sum_{e=1}^{e=k} 1_{C_{e}(x)\neq C_{e}(y)} \leq \\
    \frac{1}{k}\sum_{e=1}^{e=k} 1_{C_{e}(x)\neq C_{e}(z)}+1_{C_{e}(z)\neq C_{e}(y)} =\\
    d_C(x,z) + d_C(z,y)
\end{equation}%
Consequently, this is a pseudo-metric.

\subsection{Strict Longitudinal Distance}

The variant Strict Longitudinal Distance  considers a relevance weight based on how correct the classifier $C_{e}(x)$ is throughout its life cycle.

If the $w_e$ were independent of the $x_i$ and $x$ arguments, then the analogous computations would demonstrate $d_{SL}$, Equation \ref{e:dCC}, to also be a pseudo-metric.  However, because of $d_{SL}$'s dependence on the binary weights, $w_e(x_i)$, on the first argument, $x_i$, it fails to be a pseudo-metric when $x\in X_{\rm test}$ is an element of the training set. It does, however, satisfy the properties that $d_{SL}(x,y)\geq 0$ and $d_{SL}(x,x)=0$, so it provides some weaker notion of distance between points.  

\section{Explaining with Longitudinal Distances}\label{Explaining}
To use the proposed metrics $d_L$ and $d_{SL}$ for explaining decisions of an NN, it is necessary that the weight matrices produced at every epoch are preserved. If those have not been preserved, then it is necessary to retrain the NN preserving its weight matrices. Due to its non-deterministic nature, the instance to be explained has to be solved again by the trained classifier for an explanation to be provided.
To explain a given target instance, we propose to compute the distance between the target instance and all training instances using the longitudinal distance or the strict longitudinal distance. With the results of distances, it is then possible to determine the minimum (\emph{i.e}., shortest) distance observed between the target and training instances. Now note that multiple training instances can be equidistant to the target, and this is exactly why longitudinal distances fail the axioms for being metrics. 

\subsubsection{Definition 1}
The shortest distance observed between a given target instance and all training instances computed through longitudinal distances is defined as the \emph{explainer distance}.
\subsubsection{Definition 2}
The set of instances that are equidistant to the target instance at the \emph{explainer distance} constitute the \emph{positive explainer set}.

The \emph{positive explainer set} and the \emph{negative explainer set} are further specified in that the \emph{explainer set} computed via longitudinal instances is, by nature of equations (1) and (2), the \emph{positive explainer set}. The premise of the longitudinal distances may be reversed to compute the \emph{negative explainer set} by modifying Equation (3) as follows:
\begin{equation}
\label{e:seneg}
    \delta_e(x_i,x) = 1_{C_{e}(x_i) \neq C_{e}(x)}
\end{equation}
the $1_{C_{e}(x_i)=C_{e}(x)}$ in Equation \ref{e:seneg} is the indicator function taking the value of 1 when they are not equal and 0 otherwise. 

\subsubsection{Definition 3}
The \emph{negative explainer set} is the set of instances computed via longitudinal distances that represent negative instances in classifying a given target instance.

\subsubsection{Definition 4}
The \emph{explainer set} results from the combination of the \emph{positive explainer set} and the \emph{negative explainer set}.

The instances belonging to the \emph{explainer set} we theorize are responsible for producing the solution to the target instance and consequently can explain its solution. Once the explainer set is known, some considerations are needed. First, the \emph{explainer distance} needs to be defined based on a precision value $\mathcal{E}$. The value for $\mathcal{E}$ depends on the domain given that explanations tend to be domain-dependent. Second, a training instance may produce a direct or indirect explanation. A direct explanation would be one that does not require any further processing as in all contents of the training instance(s) suffices to explain the target instance. An indirect explanation may require further processing such as comparing whether all features of the target instance match the instances of the training instance chosen to explain the target. Third, the explainer set may include one or more training instances. When cardinality is greater than one, then a process to select the most loyal explanation may be needed. As a result of this process, it may be that no explanation is given. Fourth, if the explanation is meant to foster human trust, the explainer set may need to be interpreted. When the set is meant to produce accountability, then it has to be logical.

We note that the negative explainer set is needed to demonstrate positive and negative instances that could be potentially used to train a classifier. When used to select the best candidate for explanation, the positive explainer set suffices.

\section{Studies} \label{Studies}

\begin{figure}
\centerline{\includegraphics[width=13cm,keepaspectratio]{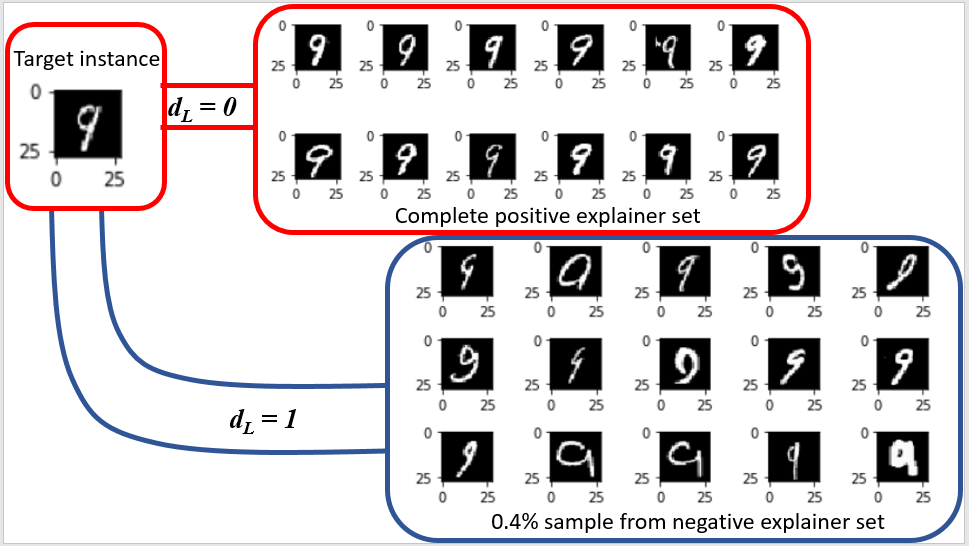}}
\caption{Image of 9 on the left is the target instance, the set circled in red is the positive explainer set with all members at $d_L$ = 0, the explainer distance for the positive set. The set circled in blue is a sample of a 3,569 (0.04\%) image instances at the negative explainer distance $d_L$ = 1}
\label{visual}
\end{figure}


\subsection{Visual Inspection}\label{visualinspection}

We used the MNIST data set consisting of 55k training instances, and 10k testing instances. We implemented a convolutional neural network design consisting of 1 convolutional layer with 28 3x3 kernels, 1 max pool layer of size 2x2 and stride of 1. The resultant layer is flattened and then connected to 1 hidden layer with 128 nodes with ReLU activation function for every node. A dropout of 20\% is introduced following the hidden layer. The resultant output is passed to the output layer with 10 nodes (each corresponding to the label) with the Softmax activation function. This entire network is trained with 55,000 28x28 MNIST images in batches of 64 over 10 epochs. This training reached an accuracy of 83.9\%\footnotemark.\footnotetext{Code and results are available https://github.com/Rosinaweber/LongitudinalDistances} 

Fig. \ref{visual} shows the images of the entire positive explainer set for one of the testing instances from MNIST labeled as a number nine. This positive explainer set was selected for illustration because its cardinality is 12 and thus we can show all the 12 images that are at distance zero from the target testing instance, measured by $d_L$. The negative explainer set contains 3,569 images. Figure set shown circled in blue is a 0.4\% sample of the complete set of images at distance 1, which included 3,569 images.

Although not a validation of accuracy, images are often presented to illustrate feature attribution methods. In Fig. \ref{visual} the negative explainer set is a random selection of the large overall set. In that sample, the types of nines are still nines but they are of a different category from those in the positive explainer set and the target instance.

\subsection{Fidelity of explanations with longitudinal distances}\label{FidelityofExplanations}
Fidelity of an explanation is a metric that aims to determine how well an explanation generated by an XAI method aligns with ground-truth \cite{amiri2020data}.This concept has been discussed from multiple perspectives (\emph{e.g}., \cite{alvarez2018towards,barr2020towards,guidotti2021evaluating,kim2018interpretability,mahajan2019preserving,man2021best,sundararajan2017axiomatic,yang2019benchmarking}), and also related to gold features \cite{ribeiro2016should}. In this study, we use the data and method from \cite{amiri2020data}. Their approach is to create artificial data sets by populating equation variables with values based on a given interval so that each instance can be explained by the equation used to create it. This way, an explanation is loyal to the data when it approximates the relationship between the variables in the original equation.  
In this section, we use a data set created from seven equations. The equations have three variables, which were randomly populated. The result of these equations, the numeric value produced when applying the values and executing the equation, is the fourth feature of the data set. The label corresponds to the equation, hence there are seven labels, $l = \{0, 1, 2, 3, 4, 5, 6\} $.
The data set consists of 503,200 instances. The experiment is as follows. 

\subsubsection {Data set and deep neural network classifier.}
We selected 403,200 instances for training and 100,000 for testing and validation. We trained a deep learning architecture with 1 hidden layer and 8 nodes. The activation for the input and hidden layers was done using ReLU functions and for the output layer using the Softmax activation function. The loss calculated was categorical cross entropy. The 403,200 training instances were trained in batches of 128 over 15 epochs and reached an accuracy of 95\% in the testing set. As required to implement the longitudinal distances, we preserved the classifiers at each epoch.
\subsubsection {Accuracy metric and hypothesis.}
 As earlier described, the data in this experiment was designed with equations that are known and are used to designate the class of the instances. This way, an XAI method produces a correct explanation if it selects to explain a target instance with a training instance that implements the same equation with which the instance being explained was built. The explanation will be given by an existing training instance, which has a label. Consequently, correctness is observed when the class label of the training instance selected to explain the target instance has the same label as the target instance being explained. With the definition of a correct explanation, we can define accuracy as the ratio of correct explanations to the total explanations produced by the proposed approach. Our hypothesis is that the proposed method using both longitudinal distances will select instances to explain the classifications produced by the deep NN architecture with average accuracy above 95\%.
 
\subsubsection {Methodology.}
With the classifier trained, we randomly selected 1,000 from the 100,000 testing instances to measure the accuracy of the proposed approach. This evaluation, as the metric describes, only seeks the selection of the class of explanations, and thus it does not use the negative explainer set. For each instance, we followed these steps:\\
\textbf{1.} Compute the positive explainer set using $d_{L}$ and $d_{SL}$. \\
\textbf{2.}  Use the label of each instance to create subsets of instances for each label.\\
\textbf{3.}  Determine the largest subset and use its label as explanation.\\ 
\textbf{4.} Assess whether label is correct.\\
\textbf{5.}  Compute average accuracy.\\
 
\paragraph{Results.}
Of the 1,000 testing instances, there were 978 and 980 correct explanations for the approach when using $d_{L}$ and $d_{SL}$, respectively. Both average accuracy levels are above 95\%, which is consistent with our hypothesis. The classifier is correct for 968 instances out of 1,000 (96.8\%), being wrong in 32 instances. Both distances led to the selection of the wrong explanation only when the classifier's predictions were wrong. However, the distances do not indicate the wrong explanations whenever the classifier is wrong. Interestingly, both distances $d_{L}$ and $d_{SL}$ were correct, respectively, in 10 and 12 instances for which the classifier was incorrect. 

\setlength{\tabcolsep}{4pt}
\begin{table}
\begin{center}
\caption{Analysis of the predictions when $d_{L}$ selected the correct explanation when the classifier was wrong. The correlation between the explainer set (1st column) and the respective number of distinct predictions (4th column) and changes (5th column) is 0.60 and 0.74}
\label{card}
\begin{tabular}{|p{0.13\textwidth}|p{0.1\textwidth}|p{0.42\textwidth}|p{0.1\textwidth}|p{0.1\textwidth}|}
\hline
Explainer set size  & Majority label set size & 15 Predictions  & Distinct predictions  & Changes\\
\hline
841	&	787	&	0, 0, 0, 4, 4, 4, 4, 4, 4, 4, 4, 4, 4, 4, 4	&	2	&	1	\\
527	&	480	&	4, 4, 4, 4, 4, 4, 4, 4, 4, 4, 4, 4, 4, 0, 4	&	2	&	2	\\
527	&	480	&	4, 4, 4, 4, 4, 4, 4, 4, 4, 4, 4, 4, 4, 0, 4	&	2	&	2	\\
321	&	321	&	4, 4, 4, 4, 4, 4, 4, 4, 4, 4, 4, 0, 4, 4, 4	&	2	&	2	\\
321	&	321	&	4, 4, 4, 4, 4, 4, 4, 4, 4, 4, 4, 0, 4, 4, 4	&	2	&	2	\\
120	&	120	&	5, 3, 3, 3, 3, 3, 3, 3, 3, 3, 1, 1, 1, 1, 1	&	2	&	2	\\
32	&	32	&	6, 6, 6, 6, 6, 6, 0, 4, 0, 4, 0, 0, 4, 4, 4	&	3	&	6	\\
29	&	29	&	6, 6, 6, 6, 6, 6, 0, 4, 4, 4, 4, 0, 4, 4, 4	&	3	&	4	\\
15	&	15	&	6, 6, 6, 6, 6, 6, 0, 4, 4, 4, 0, 0, 4, 4, 4	&	3	&	4	\\
5	&	5	&	3, 3, 3, 3, 3, 3, 3, 3, 0, 3, 3, 0, 0, 0, 0	&	2	&	3	\\
\hline
\end{tabular}
\end{center}
\end{table}
\setlength{\tabcolsep}{1.4pt}

\paragraph{Discussion.}
It is curious that the distances led to the selection of the correct explanation in instances when the classifier was wrong because the distances are based on the results of the classifier. We examined whether there was anything else unusual about those instances. We found that when $d_{L}$ was able to select correct explanations while the classifier was wrong, the cardinality of the explainer sets was much smaller than in other instances. Considering when the classifier was wrong, the average size of the explainer set was 274 instances when the correct explanation was selected in comparison to 23,239 when the explanation was not correct. We then examined the predictions of the classifier throughout the 15 epochs and computed two values, namely, the number of distinct predictions and the number of times the prediction changed. Table~\ref{card} shows the values for all instances where $d_{L}$ selected the correct explanation when the classifier was wrong. The correlation between the explainer set and the respective number of distinct predictions and changes is 0.60 and 0.74. We also computed the correlation between the explainer set and the respective number of distinct predictions and changes for when $d_{L}$ was not correct, and the results are 0.82 and 0.72. These correlations suggest that the more changes in labels the classifier makes as it is learning, the more demanding the distances become causing the explainer sets to be more efficient. There is a lot to investigate further on this finding. 

\section{Conclusions and Future Work}
In this paper, we introduced longitudinal distances to be computed between an unseen target instance classified by an NN and all training instances used to train it. The training instances at the shortest distance from the target instance constitute its explainer set. The training instances that are members of the explainer set are hypothesized to have contributed to classify the target instance. As such, they can be potentially used to explain the target instance.

Longitudinal instances are inspired by the similarity heuristics and the principle that similar problems have similar solutions. Although not demonstrated yet, instance attribution methods have the potential to bring to example-based XAI, particularly when implemented with CBR, the facet of attribution, currently missing in those approaches.

In this paper, the positive explainer set was used to select the best candidate for explanation. We did not use the negative explainer set. It is obvious that, if we are interested in accountability, as this paper described in its motivation, then we first need to demonstrate that the explainer set produces the explained decision; this is when we will use the negative explainer set. Ultimately, we also need to determine how to modify the set to change its decisions. 

New XAI and IML approaches should demonstrate how they address any criticisms. Demonstrating how they perform in presence of noise, the proportion of outliers in the explainer set, and the proportion of overlap across explainer sets for multiple instances are future work for longitudinal distances. 

\subsubsection*{Acknowledgments.} Support for the preparation of this paper to Rosina O Weber and Prateek Goel was provided by NCATS, through the Biomedical Data Translator program (NIH award 3OT2TR003448-01S1). Gideon Simpson is supported by NSF Grant no.DMS-1818716.   \bibliographystyle{splncs}
\bibliography{06062021}

\begin{thebibliography}{10}
\providecommand{\url}[1]{\texttt{#1}}
\providecommand{\urlprefix}{URL }

\bibitem{adebayo2018sanity}
Adebayo, J., Gilmer, J., Muelly, M., Goodfellow, I., Hardt, M., Kim, B.: Sanity
  checks for saliency maps. In: 32nd NeurIPS. pp. 9525--9536 (2018)

\bibitem{adler2018auditing}
Adler, P., Falk, C., Friedler, S.A., Nix, T., Rybeck, G., Scheidegger, C.,
  Smith, B., Venkatasubramanian, S.: Auditing black-box models for indirect
  influence. Knowledge and Information Systems  54(1),  95--122 (2018)

\bibitem{alvarez2018towards}
Alvarez-Melis, D., Jaakkola, T.S.: Towards robust interpretability with
  self-explaining neural networks. arXiv preprint arXiv:1806.07538  (2018)

\bibitem{amiri2020data}
Amiri, S.S., Weber, R.O., Goel, P., Brooks, O., Gandley, A., Kitchell, B.,
  Zehm, A.: Data representing ground-truth explanations to evaluate xai
  methods. arXiv preprint arXiv:2011.09892  (2020)

\bibitem{bach2015pixel}
Bach, S., Binder, A., Montavon, G., Klauschen, F., M{\"u}ller, K.R., Samek, W.:
  On pixel-wise explanations for non-linear classifier decisions by layer-wise
  relevance propagation. PloS one  10(7),  e0130140 (2015)

\bibitem{baldoni2016computational}
Baldoni, M., Baroglio, C., May, K.M., Micalizio, R., Tedeschi, S.:
  Computational accountability. In: Deep Understanding and Reasoning: A
  Challenge for Next-generation Intelligent Agents, URANIA 2016. vol. 1802, pp.
  56--62. CEUR Workshop Proceedings (2016)

\bibitem{barr2020towards}
Barr, B., Xu, K., Silva, C., Bertini, E., Reilly, R., Bruss, C.B., Wittenbach,
  J.D.: Towards ground truth explainability on tabular data. arXiv preprint
  arXiv:2007.10532  (2020)

\bibitem{barshan2020relatif}
Barshan, E., Brunet, M.E., Dziugaite, G.K.: Relatif: Identifying explanatory
  training samples via relative influence. In: International Conference on
  Artificial Intelligence and Statistics. pp. 1899--1909. PMLR (2020)

\bibitem{bhatt2019building}
Bhatt, U., Ravikumar, P., et~al.: Building human-machine trust via
  interpretability. In: AAAI Conference on Artificial Intelligence. vol.~33,
  pp. 9919--9920 (2019)

\bibitem{chen2021hydra}
Chen, Y., Li, B., Yu, H., Wu, P., Miao, C.: Hydra: Hypergradient data relevance
  analysis for interpreting deep neural networks. arXiv:2102.02515  (2021)

\bibitem{garfinkel2017toward}
Garfinkel, S., Matthews, J., Shapiro, S.S., Smith, J.M.: Toward algorithmic
  transparency and accountability. Communications of the ACM  60(9) (2017)

\bibitem{guidotti2021evaluating}
Guidotti, R.: Evaluating local explanation methods on ground truth. Artificial
  Intelligence  291,  103428 (2021)

\bibitem{NEURIPS2019_fe4b8556}
Hooker, S., Erhan, D., Kindermans, P.J., Kim, B.: A benchmark for
  interpretability methods in deep neural networks. In: Wallach, H.,
  Larochelle, H., Beygelzimer, A., d\textquotesingle Alch\'{e}-Buc, F., Fox,
  E., Garnett, R. (eds.) Advances in Neural Information Processing Systems.
  vol.~32. Curran Associates, Inc. (2019),
  \url{https://proceedings.neurips.cc/paper/2019/file/fe4b8556000d0f0cae99daa5c5c5a410-Paper.pdf}

\bibitem{kenny2019twin}
Kenny, E.M., Keane, M.T.: Twin-systems to explain artificial neural networks
  using case-based reasoning: Comparative tests of feature-weighting methods in
  ann-cbr twins for xai. In: Twenty-Eighth International Joint Conferences on
  Artificial Intelligence (IJCAI), Macao, 10-16 August 2019. pp. 2708--2715
  (2019)

\bibitem{khanna2019interpreting}
Khanna, R., Kim, B., Ghosh, J., Koyejo, S.: Interpreting black box predictions
  using fisher kernels. In: The 22nd International Conference on Artificial
  Intelligence and Statistics. pp. 3382--3390. PMLR (2019)

\bibitem{kim2018interpretability}
Kim, B., Wattenberg, M., Gilmer, J., Cai, C., Wexler, J., Viegas, F., et~al.:
  Interpretability beyond feature attribution: Quantitative testing with
  concept activation vectors (tcav). In: International conference on machine
  learning. pp. 2668--2677. PMLR (2018)

\bibitem{koh2017understanding}
Koh, P.W., Liang, P.: Understanding black-box predictions via influence
  functions. In: International Conference on Machine Learning. pp. 1885--1894.
  PMLR (2017)

\bibitem{koul2018learning}
Koul, A., Greydanus, S., Fern, A.: Learning finite state representations of
  recurrent policy networks. arXiv preprint arXiv:1811.12530  (2018)

\bibitem{kroll2017accountable}
Kroll, J.A., Huey, J., Barocas, S., Felten, E.W., Reidenberg, J.R., Robinson,
  D.G., Yu, H.: Accountable algorithms. UNIVERSITY of PENNSYLVANIA LAW REVIEW
  pp. 633--705 (2017)

\bibitem{lundberg2017unified}
Lundberg, S., Lee, S.I.: A unified approach to interpreting model predictions.
  arXiv preprint arXiv:1705.07874  (2017)

\bibitem{mahajan2019preserving}
Mahajan, D., Tan, C., Sharma, A.: Preserving causal constraints in
  counterfactual explanations for machine learning classifiers. arXiv preprint
  arXiv:1912.03277  (2019)

\bibitem{man2021best}
Man, X., Chan, E.P.: The best way to select features? comparing mda, lime, and
  shap. The Journal of Financial Data Science  3(1),  127--139 (2021)

\bibitem{ffe0793b43f842d2a50467d736a80c83}
McCloskey, M., Cohen, N.: Catastrophic interference in connectionist networks:
  The sequential learning problem. Psychology of Learning and Motivation -
  Advances in Research and Theory  24(C),  109--165 (Jan 1989)

\bibitem{mercier2020interpreting}
Mercier, D., Siddiqui, S.A., Dengel, A., Ahmed, S.: Interpreting deep models
  through the lens of data. In: 2020 International Joint Conference on Neural
  Networks (IJCNN). pp. 1--8. IEEE (2020)

\bibitem{mittelstadt2019explaining}
Mittelstadt, B., Russell, C., Wachter, S.: Explaining explanations in ai. In:
  Proceedings of the conference on fairness, accountability, and transparency.
  pp. 279--288 (2019)

\bibitem{mittelstadt2016ethics}
Mittelstadt, B.D., Allo, P., Taddeo, M., Wachter, S., Floridi, L.: The ethics
  of algorithms: Mapping the debate. Big Data \& Society  3(2),
  2053951716679679 (2016)

\bibitem{now2018annual}
NOW, A.: Annual report--new york university (2018)

\bibitem{olah2018building}
Olah, C., Satyanarayan, A., Johnson, I., Carter, S., Schubert, L., Ye, K.,
  Mordvintsev, A.: The building blocks of interpretability. Distill  3(3),  e10
  (2018)

\bibitem{ribeiro2016should}
Ribeiro, M.T., Singh, S., Guestrin, C.: " why should i trust you?" explaining
  the predictions of any classifier. In: 22nd ACM SIGKDD. pp. 1135--1144 (2016)

\bibitem{shrikumar2017learning}
Shrikumar, A., Greenside, P., Kundaje, A.: Learning important features through
  propagating activation differences. In: International Conference on Machine
  Learning. pp. 3145--3153. PMLR (2017)

\bibitem{sliwinski2019axiomatic}
Sliwinski, J., Strobel, M., Zick, Y.: Axiomatic characterization of data-driven
  influence measures for classification. In: Proceedings of the AAAI Conference
  on Artificial Intelligence. vol.~33, pp. 718--725 (2019)

\bibitem{sundararajan2017axiomatic}
Sundararajan, M., Taly, A., Yan, Q.: Axiomatic attribution for deep networks.
  In: International Conference on Machine Learning. pp. 3319--3328. PMLR (2017)

\bibitem{yang2019benchmarking}
Yang, M., Kim, B.: Benchmarking attribution methods with relative feature
  importance. arXiv preprint arXiv:1907.09701  (2019)

\bibitem{yeh2018representer}
Yeh, C.K., Kim, J.S., Yen, I.E., Ravikumar, P.: Representer point selection for
  explaining deep neural networks. arXiv preprint arXiv:1811.09720  (2018)

\end{thebibliography}

\end{document}